\documentclass[runningheads]{llncs}
\usepackage[utf8]{inputenc}

\usepackage{graphicx}
\usepackage{multirow}
\usepackage{booktabs}
\usepackage{amsmath}
\usepackage[sort, numbers]{natbib}
\usepackage{svg}
\usepackage{subcaption}
\usepackage{hyperref}

\usepackage{orcidlink}

\usepackage[misc,geometry]{ifsym}
\begin{document}
\setcitestyle{square}
\title{Improving Information Extraction on Business Documents with Specific Pre-Training Tasks}
\titlerunning{Improving Documents Information Extraction with Specific Pre-Training}
%
\author{\mbox{Thibault Douzon\inst{1,2} \orcidlink{0009-0001-1649-5373}} \and \mbox{Stefan Duffner\inst{1} \orcidlink{0000-0003-0374-3814}} \and \mbox{Christophe Garcia\inst{1} \orcidlink{0000-0001-7997-9837}} \and \mbox{Jérémy Espinas\inst{2}}}

\authorrunning{T. Douzon et al.}

\institute{INSA Lyon, LIRIS, Lyon, France \\ \email{firstname.lastname@insa-lyon.fr}
\and Esker, Lyon, France \\ \email{firstname.lastname@esker.com}
}
%
%
\maketitle              
\begin{abstract}
Transformer-based Language Models are widely used in Natural Language Processing related tasks. Thanks to their pre-training, they have been successfully adapted to Information Extraction in business documents. However, most pre-training tasks proposed in the literature for business documents are too generic and not sufficient to learn more complex structures.
In this paper, we use
LayoutLM, a language model pre-trained on a collection of business documents,
and introduce two new pre-training tasks that further improve its capacity to extract relevant information. 
The first is aimed at better understanding the complex layout of documents, and the second focuses on  numeric values and their order of magnitude. 
These tasks force the model to learn better-contextualized representations of the scanned documents.
We further introduce a new post-processing algorithm to decode \texttt{BIESO} tags in Information Extraction that performs better with complex entities.
Our method significantly improves extraction performance on both public (from 93.88 to 95.50 F1 score) and private (from 84.35 to 84.84 F1 score) datasets composed of expense receipts, invoices, and purchase orders.
%

\keywords{Business Documents \and Document Understanding \and Information Extraction \and Pre-Training \and \texttt{BIESO} Decoding \and Transformer}
\end{abstract}
\section{Introduction}

Business documents are paper-sized files containing useful information about interactions between companies. They may take the form of invoices, purchase orders, various reports, and agreements. The exact layout of a document depends on the issuer, but the contained information is conventionally structured. For example invoices and purchase orders share the same header, table, footer structure that almost all issuers have adopted.
Because such documents trace every transaction made by companies, they are the key to business process automation.
With the emergence of modern resource planning systems, accurate Information Extraction (IE) has become one of the core problems of Document Intelligence.

Initially, information extraction was done by human operators, but software solutions have been developed since the early days of document analysis to tackle the problem. 
Their intent was to ease the work of human operators with hard-coded extraction rules. Unfortunately, these rules needed to be adapted for each and every layout of documents. 
This limitation has led to the rise of Machine Learning (ML) models for automatic document IE. 

First ML approaches relied on Optical Character Recognition (OCR) systems to provide the textual content of the document. 
This transition from image to text allowed for standard Natural Language Processing (NLP) methods to be applied by adopting a Sequence Labeling problem. The amount of labeled data necessary to train accurate NLP models has always been a problem. Business documents are inherently private which strongly limits the quantity of publicly available data. 
Thus, only companies selling business process automation software are able to collect larger amounts of such data. 
Moreover, they often rely on their customer to implicitly label the documents.

\begin{figure}[b!]
  \begin{minipage}[c]{0.45\textwidth}
    \includegraphics[trim=0 200 0 80 ,scale=0.3]{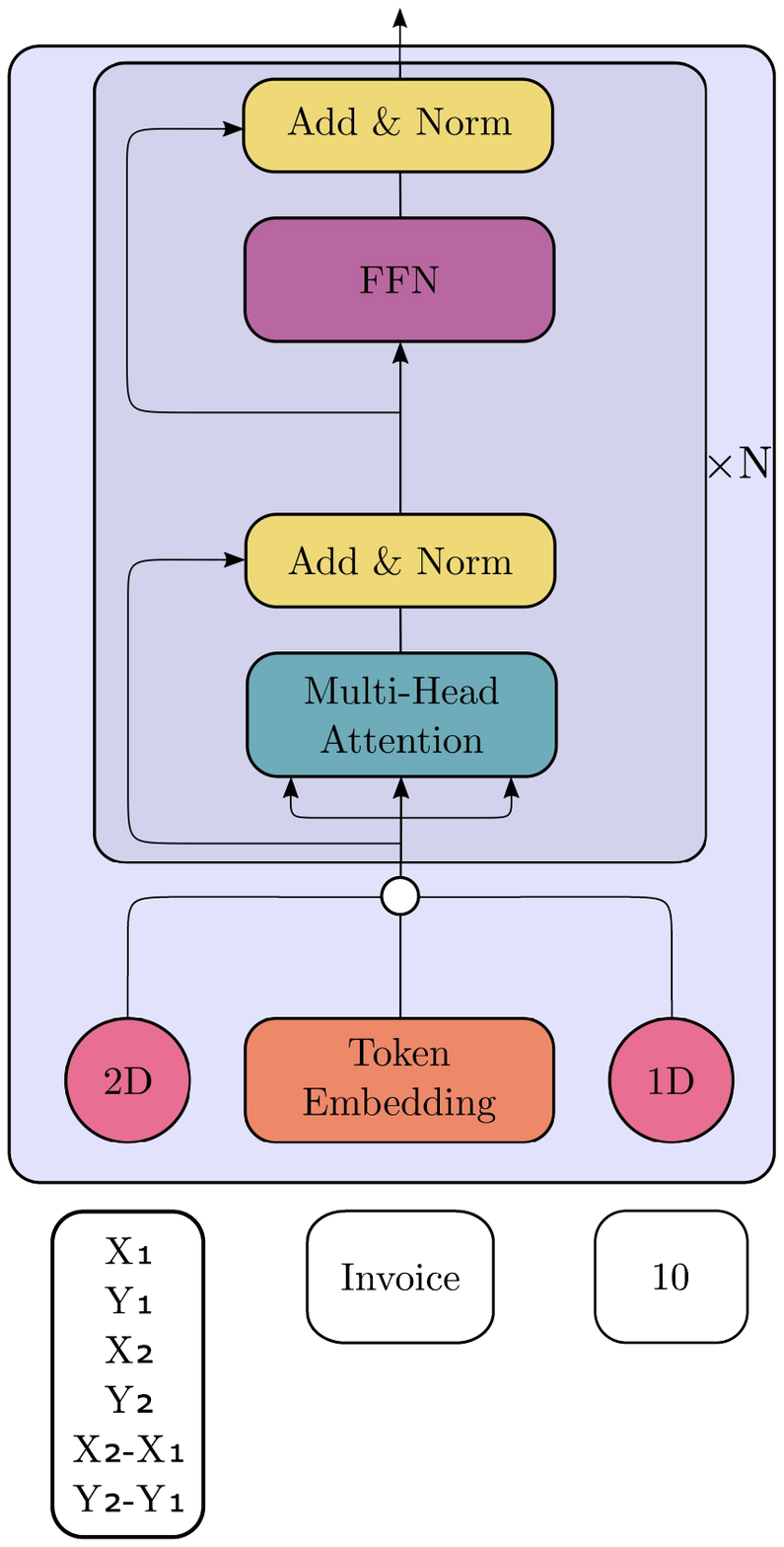}
  \end{minipage}\hfill
  \begin{minipage}[c]{0.55\textwidth}
    \caption{LayoutLM architecture.\\ Token embeddings are enriched with 1D positional encoding and 2D spatial encoding specific to this architecture. The number of blocks $N$ varies from 12, for the base model, to 24, for the large one.}
    \label{fig:layoutlm}
  \end{minipage}
\end{figure}
Most recent proposals often include a pre-training step, where the model is trained on a pretext task. 
Those pretext tasks are self-supervised problems that teach the model many useful ``skills'' for manipulating the data. Usually, these tasks are as broad as possible, teaching the model common sense about language, grammar, and global structure. 

In this work, we focus on LayoutLM~\citep{xu_layoutlm_2020}, a pre-trained Transformer that is specialized in business documents. As shown in \figurename~\ref{fig:layoutlm}, it reuses the same Transformer layer with multi-head attention with the addition of a 2D positional encoding. Its larger version achieved state-of-the-art performance in both document classification and information extraction. However, the required hardware to train it can be repelling. 
In this paper, we propose new pre-training tasks specific to business documents that will provide additional skills to the model. We also propose a new decoding post-processing algorithm that prevents many errors made by the model due to ambiguities. Combined, our contributions\footnote{\scriptsize Code available here: \url{https://github.com/thibaultdouzon/business-document-pre-training}} allow for the base LayoutLM model to perform on par with the large version.


%

\section{Related Work}

\subsection{Information Extraction}
Rule-based approaches~\citep{li_regular_2008} have been supplanted by Deep Learning models in the last decade. Document IE first capitalized on the state of the art in Named Entity Recognition for NLP~\citep{lample_neural_2016}. Recurrent Neural Networks with Long-Short Term Memories were first used to encode documents at a word level~\citep{palm_cloudscan_2017, sage_recurrent_2019}, allowing a simple classifier to predict each word's associated label. Instead of a softmax and cross-entropy loss, a Conditional Random Field~\citep{sutton_introduction_2010} model has been used in addition to \texttt{BIESO} tags. 
Other architectures have also been proposed to better adapt to the specificity of the document. For example, graphs~\citep{carneiro_invoice_2019,liu_graph_2019,yu_pick_2020,gal_cardinal_2020} and convolutions over a grid~\citep{katti_chargrid_2018,denk_bertgrid_2019,lin_vibertgrid_2021} constrained the model based on the words' positional information. 
Because most architectures relied on textual representations, they benefited from pre-trained word embeddings like Word2Vec~\citep{mikolov_efficient_2013} or GloVe~\citep{pennington_glove:_2014}.

With the emergence of Transformers~\citep{vaswani_attention_2017} and text encoders like BERT~\citep{devlin_bert_2019}, attention-based document analysis models~\citep{xu_layoutlm_2020,garncarek_lambert_2020,xu_layoutlmv2_2021} evolved quickly, which resulted in a large improvement of state-of-the-art performance. In line with~\citep{katti_chargrid_2018} which included both textual and visual representations, multi-modal Transformers~\citep{xu_layoutlmv2_2021,lin_vibertgrid_2021} superseded conventional textual models.

In parallel to the rise of Transformers, end-to-end IE models tried to reduce the labeling cost. 
First using RNNs with attention layers~\citep{palm_end--end_2019,sage_end--end_2020}, then shifting to Transformers~\citep{powalski_going_nodate}. Adopting at the same time the Question Answering~\citep{gardner_question_2019} (QA) format, instead of the usual Sequence Labeling, provided more flexibility on the predicted labels.

\subsection{Pre-Training}

Semi-supervised training and pre-trained models were popularised in NLP with enriched word embeddings~\citep{peters_deep_2018,mikolov_efficient_2013,pennington_glove:_2014}.
With the emergence of Transformers, large pre-trained models have been proposed~\citep{wolf_huggingfaces_2020}. Thanks to their pre-training, they can efficiently adapt to various tasks~\citep{wang_glue_2019,wang_superglue_2020} and data types. In general, these models are pre-trained on large unlabeled datasets in a self-supervised manner. This self-supervision removes parts of the burden of data labeling~\citep{sage_data-efficient_2021} and leverages the huge quantities of available data. 

A wide variety of pre-training tasks have been proposed. General-purpose tasks aiming at learning the language and grammar were used first. Auto-regressive tasks~\citep{radford_improving_2018} and Masked Language Modeling~\citep{devlin_bert_2019} are still frequently used in new pre-trained models as they have proven to be effective in most situations. In addition to incremental improvements~\citep{raffel_exploring_2020,yang_xlnet_2020}, some new pre-training tasks were designed to align representations of multi-modal inputs~\citep{xu_layoutlmv2_2021,powalski_going_nodate}.









\section{Models}
\subsection{Architecture}
We used the well-established LayoutLM architecture~\citep{xu_layoutlm_2020} which itself is based on BERT Transformer~\citep{devlin_bert_2019}. 
More specifically, we chose the base model\footnote{\scriptsize Pre-trained weights available here: \url{https://huggingface.co/microsoft/layoutlm-base-uncased}} with 12 layers and 512 dimensions for token embeddings. This model is computationally much more efficient compared to the larger version while still giving very good performance.


Transformer models work on tokens that are in between characters and words. LayoutLM and BERT both use the WordPiece algorithm. We use the same tokenizer as LayoutLM in order to compare our performance with the base LayoutLM model. It uses a vocabulary size of $30 000$, and we limit the sequence length to 512 tokens, including the special tokens \texttt{[CLS]} and \texttt{[SEP]}. This limitation due to GPU memory consumption of self-attention operations often forces us to cut documents in multiple pieces of 512 tokens and process them separately.

Contrary to RNNs, all positions in the sequence are equivalent in a Transformer model. To provide information about position inside the sequence, a linear positional encoding~\citep{vaswani_attention_2017} is added for each token. Then LayoutLM adapted this positional encoding to a 2D version that can represent the positions of words on a page.

For both pre-training tasks and fine-tuning, we use a simple dense layer to map each token's final internal representation to the dimension of the prediction space. A softmax layer is applied to produce the final model confidence scores. For training, the cross-entropy loss is used on the model confidence scores.

\subsection{ConfOpt Post-Processing}
We model the Information Extraction task as sequence tagging on tokens. Predictions are done at the token level and then aggregated by a lightweight post-processing  step to give the model's final prediction. 
In all experiments, we use \texttt{BIESO} tagging.
That is, each field to extract is composed of a sequence of target tags of the following types: \texttt{B} for the beginning of the entity, \texttt{I} for inside, \texttt{E} for its end, or otherwise \texttt{S} for a single token entity. \texttt{O} is used for any token that is outside any target label. 
\texttt{BIESO} is widely used in IE as it provides structure to the target sequence that helps the model. 

Instead of the trivial post-processing which consists of simply following the maximum confidence of the model, we decided to decode a model's prediction by solving a basic optimization problem. We will refer to this method as ConfOpt in the remaining of the paper.
The predicted sequence for a target label is the sequence that maximizes model confidence over the whole input sequence. There is a constraint to decode a prediction: it must match the following regular pattern: \texttt{(BI*E)\,$\vert$\,S} where \texttt{*} denotes zero or many occurrences and \texttt{$\vert$} denotes an alternative.

This optimisation problem can be solved with a dynamic programming approach. The model's predictions for one target label can represented as a $4 \times N$ dimensional matrix where $N$ is the sequence length and 4 comes from the 4 tags \texttt{B,I,E,S}. By noting $C_{T,0}$ the model's confidence in \texttt{T} tag at position 0 and $P_{T,i}$ the best prediction confidence ending at token $i$ with tag \texttt{T}, the objective is to determine $S~=~\underset{T \in \{E,S\}}{\underset{0 \leq i < N}{\max}} P_{T,i}$ where 
$$
\begin{aligned}
P_{B,i} = C_{B,i} ~;~  
P_{I,i} = C_{I,i} + \max
\begin{cases}
P_{B,i-1}\\
P_{I,i-1}\\
\end{cases}
\\
P_{S,i} = C_{S,i} ~;~
P_{E,i} = C_{E,i} + \max
\begin{cases}
P_{B,i-1}\\
P_{I,i-1}\\
\end{cases}
\end{aligned}
$$

One drawback of this post-processing is dealing with no prediction and non-unique predictions. It can be solved with an empirically determined threshold below which no predictions are made. Though in this paper this is not further studied because fields are mandatory in a document and always unique.

\section{Pre-training}
Transformer models provide great performance when first pre-trained on pretext tasks on very large unlabelled datasets. This pre-training is most of the time done in a self-supervised manner in order to avoid the labeling cost. LayoutLM uses Masked Visual-Language Modeling~\citep{xu_layoutlm_2020} which is adapted from BERT's Masked Language Modeling~\citep{devlin_bert_2019}. It teaches the model how text and documents are formed at a token level. In practice, at each training step, 15\% of the tokens are randomly chosen and replaced by either a \texttt{[MASK]} token, a random token, or not replaced at all. The model tries to guess which token is the most probable right replacement at those positions.

For all pre-training tasks when a document is too long to be processed at once, we randomly select a continuous span of words of maximum size and provide it to the model instead. We expect the model to learn useful features on various parts of documents thanks to the long training. 
For very short documents, the input is padded to the maximum size.

We introduce two new specific pre-training tasks in addition to Masked Visual-Language Modeling (MVLM). The first one, Numeric Ordering teaches the model how to compare and order numbers. The second one, Layout Inclusion focuses on words in the 2D plane and their relative positioning. 
We chose to avoid regression tasks, even though their implementation would have been simpler. For example, simply removing the 2D positioning of some tokens, and asking the model to predict tokens' position is an alternative to what we propose. But this does not behave well for a token that could appear either at the top or the bottom of the document: the model would learn its mean position -- the middle -- where the token would never appear. 
In the following, we will describe the two pre-training tasks in detail.


\subsection{Numeric Ordering Task}
\begin{figure}[t!]
  \centering
  \includegraphics[trim=110 100 100 80, scale=0.5]{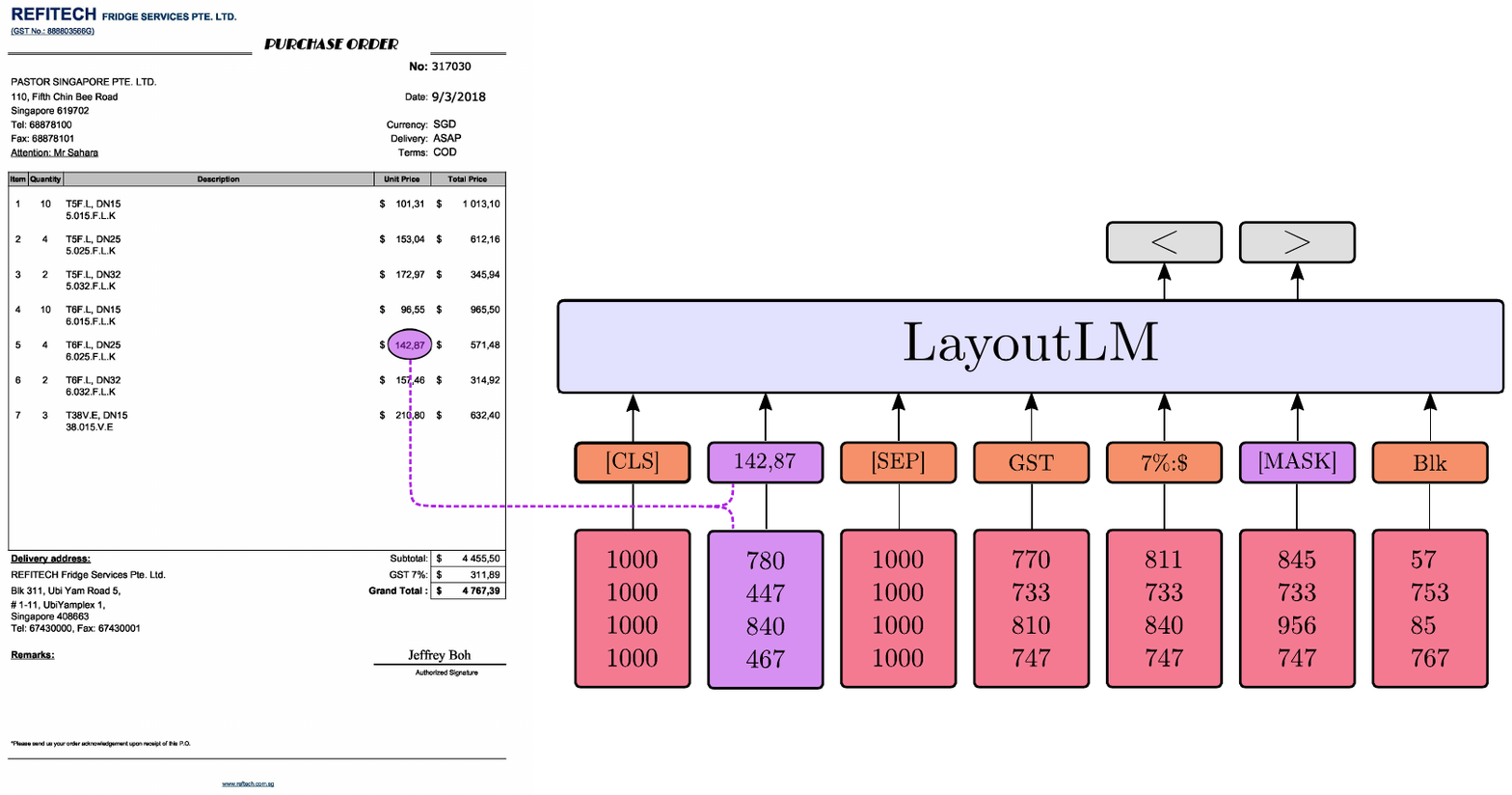}
  \caption{A pre-training example with Numeric Ordering task. A random token containing a number is selected, then the target is to predict whether other numbers are smaller or bigger. Some random noise can be added by masking tokens' textual or spatial representations. Only a small part of the document's input is represented in this illustration.}
  \label{fig:numeric_ordering}
\end{figure}

Numeric Ordering (NO) focuses on numeric figures in the document and their relative values. Contrary to MLM which only relies on self-supervised data, NO relies on a handcrafted number parser to find and parse all numbers that appear in a document. Because business documents are mostly made of decimal numbers written with digits, we ignore those written out in words.
The numeric value of each token is determined by parsing beforehand each word in the document, looking for numbers and ignoring irrelevant characters. 

As shown in \figurename~\ref{fig:numeric_ordering}, the model must predict for every numeric figure in the document if its parsed value is smaller, equal or greater than a randomly selected number among the document. The loss is only computed on tokens starting a new word, but tokens continuing a word are important to determine the value represented by a word. 

We want the model not only to reason on the textual features, but also on the spatial context surrounding each figure in the document. Therefore, we randomly mask the textual representations of 15\% of the numbers in the document and replace them with the \texttt{[MASK]} token as shown in \figurename~\ref{fig:numeric_ordering}. For the same reason, we also mask the spatial encoding of 15\% of the numbers and make sure both text and position are not masked at the same time. All masked positions are replaced with (1000, 1000, 1000, 1000).

\subsection{Layout Inclusion Task}
\begin{figure}[t!]
  \centering
  \includegraphics[trim=110 80 100 80, scale=0.5]{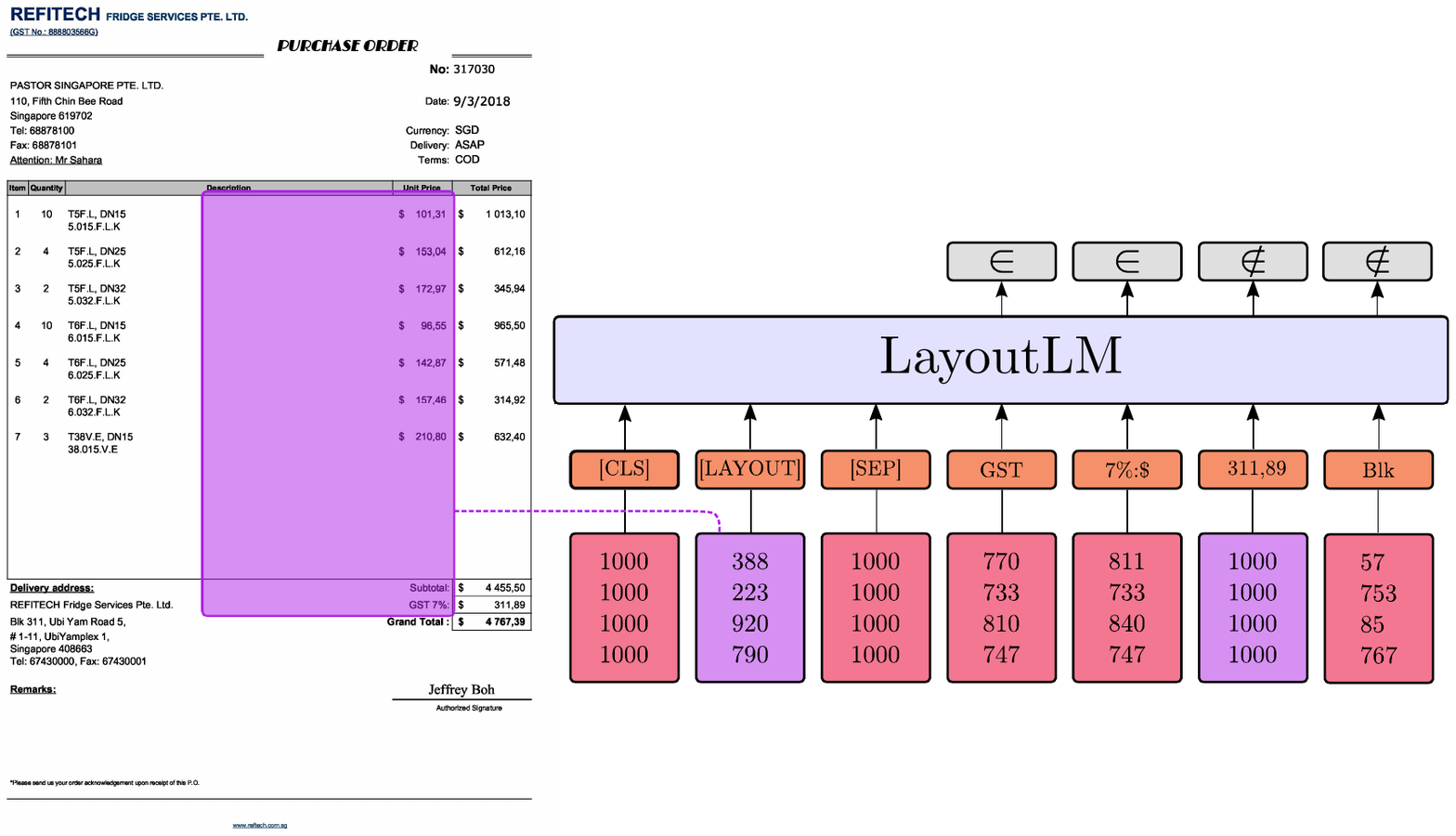}
  \caption{A pre-training example with Layout Ordering task. Coordinates of the purple rectangle are drawn uniformly. Random noise is added by masking the 2D position of some tokens. Only a small part of the document is represented.}
  \label{fig:layout_inclusion}
\end{figure}
We introduce another pre-training task focusing on the 2D positional encoding, which we called Layout Inclusion (LI). Its purpose is to provide a better understanding of document layouts and complex structures. In fact, most business documents, including invoices, purchase orders, and expense receipts, contain tables where the meaning of tokens is mostly driven by their position relative to headers. 

As shown in \figurename~\ref{fig:layout_inclusion}, Layout Inclusion is formatted like a question answering prompt: a question followed by the content of the document. The question is simply a special token \texttt{[LAYOUT]} positioned at random coordinates $(x_1, y_1, x_2, y_2)$. The model must then classify every token in the document into 2 groups: either \texttt{inside} or \texttt{outside} of the question token. 
More precisely, the target answer is whether the middle point of a document token is inside or outside the rectangle described by the coordinates of the question.

Again, the objective is for the model to not only reason on the 2D positions of tokens but also use their textual embedding. In order to force the model to use both representations, we randomly replace 15\% of documents token positions with (1000, 1000, 1000, 1000). In case of a random position replacement, the target value is still computed based on the real position of the token, and the model must make its prediction based on the token's text and the neighboring tokens using the classical 1D positional encoding.

\section{Datasets}

\begin{figure}[ht]
	\centering
    \begin{subfigure}[b]{0.384\textwidth}
		\centering
		\frame{\includegraphics[trim=0 0 110 100, width=\columnwidth]{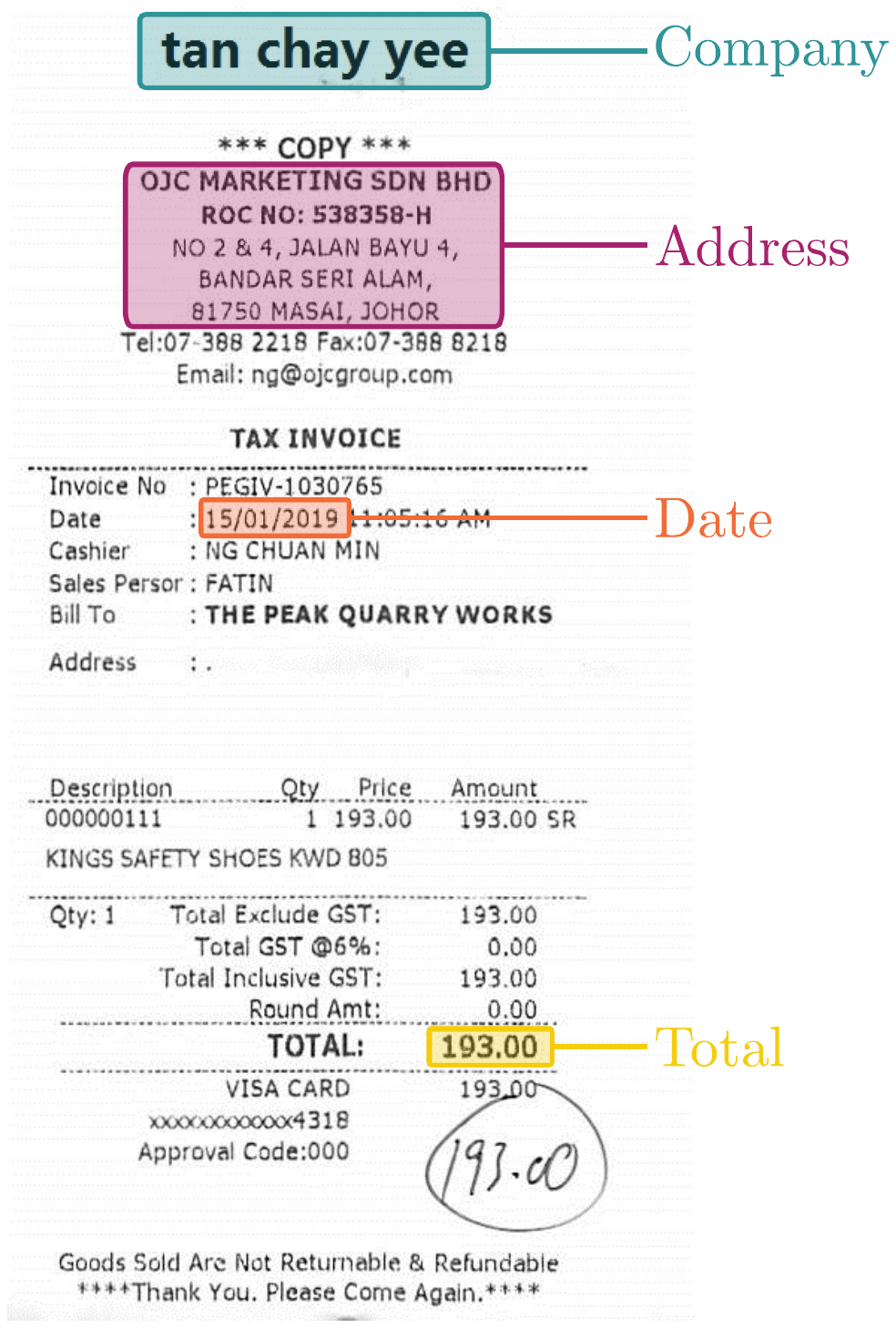}}
        \caption{Receipt from SROIE.}
		\label{fig:sample_doc_SROIE}
    \end{subfigure}
	\qquad
	\begin{subfigure}[b]{0.55\textwidth}
		\centering
		\frame{\includegraphics[trim= -5 0 0 200, width=\columnwidth]{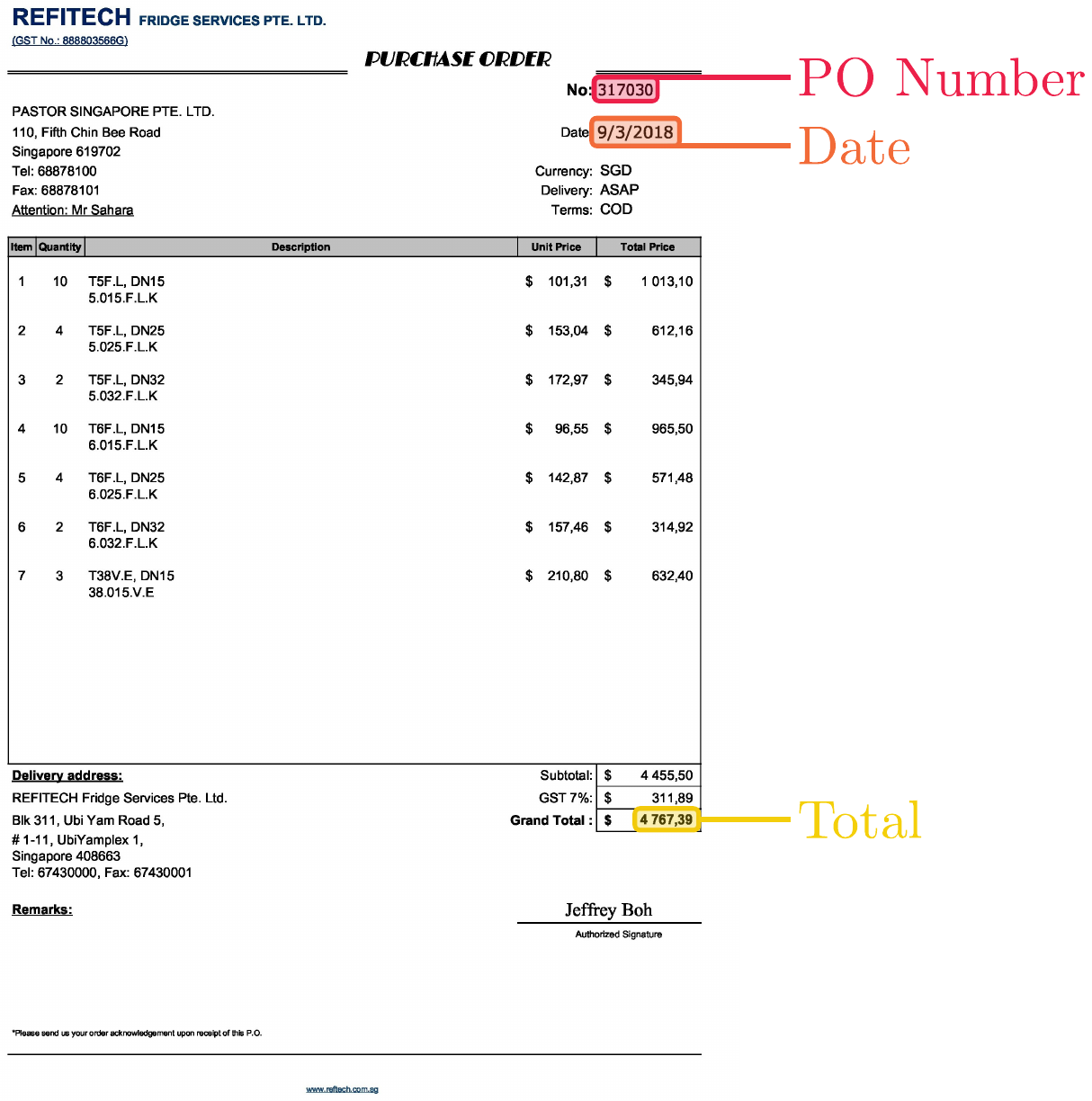}} 
        \caption{Purchase Order from BDC-PO.}
		\label{fig:sample_doc_PO-51k}
    \end{subfigure}
	\caption{A document sample for each training dataset annotated with the expected predictions. For BDC-PO, we replaced the document with a fictive one due to privacy reasons.}
	\label{fig:sample_docs}
\end{figure}

We used 2 different collections of documents to build 3 datasets for training and evaluation as described in the following. They all contain business documents: invoices and purchase orders for the private collection and expense receipts for the public one. The largest dataset used for pre-training isn't  labeled, document samples with their target fields for the others datasets are shown in \figurename~\ref{fig:sample_docs}. 
\vspace{-1cm}

\subsection{Business Documents Collection}
The Business Documents Collection (BDC) is a large private dataset composed of 100k invoices and 300k purchase orders. Those real documents were submitted and processed on a commercial document automation solution in the last 3 years. It contains English-only documents divided into 70000 different issuers. All documents sharing the same issuer usually use the same information template. Therefore, we limited the maximum number of documents of the same issuer to 50. It is important to keep the number of similar layouts in the collection low and the variety of examples high. We used this collection for pre-training language models on business documents that are closer to our final objective than RVL-CDIP~\citep{lewis_building_2006}. 

Textual and positional information have been extracted using a commercial OCR system. It achieves excellent accuracy on properly scanned documents and provides accurate word positions. We also use the provided read order to determine the order of tokens when feeding the network. This order determines the 1D positional encoding given to each token that complements the 2D positional encoding.

Because we only used this collection for pre-training models on self-supervised tasks, most documents do not have extraction labeling. Only a subset composed of purchase orders is labeled for the IE task.

\subsection{Business Documents Collection -- Purchase Orders}

We selected a subset of the Business Documents Collection to build a labeled dataset of English purchase orders called BDC-PO. It contains almost 9000 different issuers split into training, validation, and test set. In order to not introduce bias for models pre-trained on the BDC, we removed from BDC all documents emitted by a supplier contained in the test set. This means that document layouts contained in the test set have never been seen before by the model at pre-training or training time.

Long purchase orders are rare but can sometimes be longer than 20 pages. If we wanted to train models and make predictions on such documents, we would have to evaluate the model on dozens of inputs for one document. Instead, we chose to limit documents to one page and crop the remaining. It only concerns roughly 25\% of the dataset and sometimes impacts the prediction because labels are missing from the input.

The extraction task consists of 3 fields: document number, delivery date, and total amount. Those fields were chosen because they are mandatory for most customers and thus are well labeled at the word level by the end-user. We controlled the labeling quality at the issuer level and rejected from the dataset some issuers with undesirable labeling practices.

\subsection{ICDAR 2019 -- Scanned Receipts}
We also trained and evaluated our model on the public Scanned Receipts OCR and Information Extraction~\citep{huang_icdar2019_2019} (SROIE) dataset that was published for ICDAR 2019. We focus on the third task which consists in extracting information from the documents. SROIE contains Malaysian receipts split into 626 train and 347 test documents. Unfortunately, we do not have control over the composition of the test set, and most of the test layouts also occur in the training set.

We used the OCR text provided with the dataset instead of using our own OCR system. As others have pointed out~\citep{xu_layoutlm_2020}, it contains numerous little errors that negatively affect the final performance. 
For a fair comparison with the leaderboard, we manually fixed them such that the expected string appears in the input, at least. These fixes mostly concern addresses and company names. It almost exclusively involves fixing errors related to white-spaces and commas.

\section{Experiments}
All experiments were performed on a single machine equipped with two Nvidia RTX A6000 with 48Go of video memory each. This allowed us to boost the batch size up to 32 per device on a base transformer model. To further increase the batch size, we also aggregated 6 batches together before propagating the gradient for a total batch size of 192. We used the Adam optimizer with a learning rate of $1e-5$ and 200 linear warm-up steps as it improved our model's convergence. We used 1500 training steps for SROIE and 3000 steps for BDC-PO.
Finally, we ran each fine-tuning 10 times in each setup to get a precise idea of the performance of the models and the variability of the results. 
For the different pre-training scenarios, we performed only two runs and the best model was kept.

\subsection{Post-Processing}
This first set of experiments aims at comparing the post-processing used to decode the sequence produced by the model. We want to determine whether our proposed ConfOpt algorithm is competitive with other decoding methods. We decided to use the LayoutLM base model and compare the proposed ConfOpt against two other decoding algorithms as shown in Table ~\ref{tab:postprocessing_res}. 

We named Ad-Hoc the basic decoding using the label with maximal confidence for each token.
When decoding with this method, a \texttt{B} tag starts a new entity, a \texttt{I} tag continues the previous entity, a \texttt{E} closes the previous entity, and a \texttt{S} tag produces a new entity and closes it right away. Ad-Hoc and ConfOpt use the same model weights in this experiment as they do not introduce any trainable parameters. 

The second decoding algorithm uses a Conditional Random Field (CRF)~\citep{sutton_introduction_2010, lample_neural_2016} that processes LayoutLM's predictions. In this particular case, we did not use the classical cross-entropy loss but the score provided by the CRF layer. 
Because the CRF required specific training and did not optimize the same loss, its weights are different from the two other post-processing methods.

\vspace{-4mm}
\begin{table}[h!]
    \centering
    \begin{tabular}{c | c | c}
        & \multicolumn{2}{c}{Fine Tuning (F1 score)} \\
        Post Processing & SROIE & BDC-PO \\ \hline
        Ad-Hoc & $93.88 \pm 0.59$ & $84.35 \pm 0.12$ \\
        CRF    & $94.01 \pm 0.55$ & $84.40 \pm 0.16$ \\
        ConfOpt & $\mathbf{94.94 \pm 0.38}$ & $\mathbf{84.57 \pm 0.10}$ \\
    \end{tabular}
    \caption{Performance comparison on SROIE and BDC-PO between multiple post-processing algorithm. Score is computed on the exact match between the prediction and the target string.}
    \label{tab:postprocessing_res}
\end{table}
\vspace{-4mm}

We evaluated these algorithms on both SROIE and BDC-PO. The results in \tablename~\ref{tab:postprocessing_res} show a tiny improvement using a CRF instead of the Ad-Hoc post-processing (0.13 and 0.05 F1 points) but those differences are always within one standard deviation range. We would need more evidence to conclude on the effect of adding a CRF layer for the post-processing. 

On both datasets, using ConfOpt significantly increases performance (1.06 and 0.22 F1 points) compared to the Ad-Hoc post-processing, even though the model is strictly identical. In light of these results, we decided to use the ConfOpt for the next experiment.

\subsection{Business Document-Specific Pre-training}
We conducted another set of experiments in order to study the effects of the new business data-specific pre-training tasks on the model performance. At the same time, we controlled the performance gap obtained by pre-training with the basic MVLM task on the same new dataset. Both comparisons are insightful to decide whether it is useful to pre-train on clients' data and/or with data-specific pre-training tasks.

For the pre-training part, we always initialize the model's weights with the base version~\citep{xu_layoutlm_2020}. We pre-train models for 20 epochs on 80\% of BDC. 
When using multiple pre-training tasks at the same time, we chose to provide batches of single tasks to the model. Gradient aggregation over multiple batches helps smoothing the update between different tasks. We pre-trained 2 models on the BDC, one with MVLM only and another with MVLM+NO+LI.

\vspace{-3mm}
\begin{table}[!h]
    \centering
    \begin{tabular}{c c | c | c c c}
        \multicolumn{2}{c|}{Pre Training} & \multicolumn{1}{c}{} & \multicolumn{3}{|c}{Accuracy per field} \\
        Task(s) & Dataset & F1 Score & PO Number & \hspace*{0.8mm} Total \hspace*{1mm} & \hspace*{1.2mm} Date \hspace*{1.2mm}  \\ \hline
        MVLM & RVL-CDIP      & $84.57 \pm 0.10$ & $89.98$ & $89.10$ & $93.59$\\
        MVLM & BDC           & $84.77 \pm 0.12$ & $90.61$ & $89.33$ & $93.59$ \\
        MVLM+NO+LI & BDC     & $\mathbf{84.84 \pm 0.08}$ & $\mathbf{90.71}$ & $\mathbf{89.36}$ & $\mathbf{93.83}$ \\
    \end{tabular}
    \caption{Model performance when fine-tuning on BDC-PO}
    \label{tab:pretrain_res_bdcpo}
\end{table}
\vspace{-1cm}
\begin{table}[!h]
    \centering
    \addtolength{\leftskip} {-2cm}
    \addtolength{\rightskip}{-2cm}
    \begin{tabular}{c | c c | c | c c c c}
        Architecture & \multicolumn{2}{c|}{Pre Training} & \multicolumn{1}{c}{} & \multicolumn{4}{|c}{Accuracy per field} \\ 
        & Task(s) & Dataset & F1 Score & Company & Address & \hspace*{1mm} Total \hspace*{0.8mm} & \hspace*{0.8mm} Date \hspace*{1mm} \\ \hline
        LayoutLM base * & MVLM & RVL-CDIP  & $94.94 \pm 0.38$ & $92.91$ & $90.81$ & $89.25$ & $99.48$ \\
        LayoutLM base * & MVLM & BDC       & $95.18 \pm 0.23$ & $\mathbf{93.72}$ & $91.00$ & $89.48$ & $\mathbf{99.68}$ \\
        LayoutLM base * & MVLM+NO+LI & BDC & $\mathbf{95.50 \pm 0.22}$ & $93.60$ & $\mathbf{91.41}$ & $\mathbf{90.89}$ & $99.57$ \\ \hline \hline
        LayoutLM base~\cite{xu_layoutlm_2020} & MVLM & RVL-CDIP & $94.38$ & \multicolumn{4}{c}{} \\
        LayoutLM large~\cite{xu_layoutlm_2020} & MVLM & RVL-CDIP & $95.24$ & \multicolumn{4}{c}{} \\
        LayoutLMv2 large~\cite{xu_layoutlmv2_2021} & MVLM+TIA+TIM & RVL-CDIP & $\mathbf{97.81}$ & \multicolumn{4}{c}{} \\
    \end{tabular}
    \caption{Model performance when fine-tuning on SROIE. Models name ending with a * are our contribution. The second part contains published scores of the original LayoutLM and LayoutLMv2 as a comparison.}
    \label{tab:pretrain_res_sroie}
\end{table}

\vspace{-3mm}

We evaluated each pre-trained model on both datasets, the results are available in \tablename~\ref{tab:pretrain_res_bdcpo} for BDC-PO and \tablename~\ref{tab:pretrain_res_sroie} for SROIE . Each cell contains the means of 10 runs with different seeds and the standard deviation is provided for the F1 score. There are a few interesting things to notice.

The first important remark is the importance of the pre-training dataset. Pre-training on BDC significantly improves performance on both SROIE and BDC-PO, even though the pretext training task is the same as what was used for LayoutLM. 
BDC is more homogeneous and focuses on invoices and purchase orders. Contrary to our expectations, we observe a greater improvement on SROIE than on BDC-PO (0.24 vs 0.2 F1 points). But the overall improvement by using BDC can be explained because RVL-CDIP contains a broader panel of document types and is not specialized like BDC. Even though BDC does not contain expense receipts, its global structure is similar to invoices.

Next, we can compare the pre-training tasks. Introducing Numeric Ordering (NO) and Layout Inclusion (LI) tasks also improves the performance over the previously pre-trained model. We observe a 0.32 F1 point improvement on SROIE but only 0.07 on BDC-PO. We suspect the small improvement introduced by the new tasks can be explained because most useful skills to process purchase orders were learned by pre-training on such documents. The new pre-training tasks  help more for generalizing on new types of documents.

We also can look at the results on a field per-field basis. We observe that using BDC over RVL-CDIP improved the recognition of all fields except for the dates in BDC-PO. If introducing new training tasks did not improve all fields, we notice that some fields were greatly enhanced like the total amount in SROIE (1.41 F1 points difference). We expected to observe a greater improvement in the total field with the new pre-training tasks. But it does not seem to improve performance much on BDC-PO's total.  

Finally it is interesting to compare on \tablename~\ref{tab:pretrain_res_sroie} our results with the published scores of LayoutLM and LayoutLMv2. Our pre-trained model with NO and LI tasks performs better than LayoutLM large which contains 3 times more parameters. However, LayoutLMv2 -- which uses both textual and visual information -- performance level is still unreachable for a textual-only model.

\section{Conclusion}

In this work, we showed significant improvements are accessible without introducing more trainable parameters and computational complexity. Only using the base transformer architecture, we achieved a performance that is comparable to the large version which contains 3 times more parameters. Pre-trained models can be further specialized through in-domain datasets and specific pretext training tasks. We demonstrated that by introducing a new collection of business documents and training tasks focusing on documents' layout and number understanding. We showed that performance improvements can be imputed to both pre-training tasks (Numeric Ordering and Layout Inclusion) and new pre-training dataset.

In the future, we will investigate on IE as a Question Answering problem. It has already been proposed in the past~\citep{gardner_question_2019} as an alternative to Sequence Labeling when fine-tuning models. It should improve the model's generalization capabilities and enable few-shot learning. But nowadays all models are pre-trained, and we would like to study the impact on generalization of a QA-only pre-training. 

%
%
%
%
\bibliographystyle{plainnat}
\bibliography{main}

\end{document}